\documentclass{article}

\usepackage{microtype}
\usepackage{graphicx}
\usepackage{subcaption}
\usepackage{booktabs} 
\usepackage{url}
\usepackage{makecell}
\usepackage{multirow}
\usepackage{hyperref}
\usepackage{etoolbox}

\AtBeginEnvironment{table}{
  \setlength{\textfloatsep}{3pt}
}

\usepackage[preprint]{icml2026}

\usepackage{amsmath}
\usepackage{amssymb}
\usepackage{mathtools}
\usepackage{amsthm}

\usepackage[capitalize,noabbrev]{cleveref}

\theoremstyle{plain}

\theoremstyle{definition}

\theoremstyle{remark}

\usepackage[textsize=tiny]{todonotes}

\icmltitlerunning{Between the Layers Lies the Truth}

\begin{document}

\twocolumn[
	\icmltitle{Between the Layers Lies the Truth: \\ Uncertainty Estimation in LLMs Using Intra-Layer Local Information Scores}
		
		
		
	\icmlsetsymbol{equal}{*}
		
	\begin{icmlauthorlist}
		\icmlauthor{Zvi N. Badash}{dds_technion}
		\icmlauthor{Yonatan Belinkov}{cs_technion}
		\icmlauthor{Moti Freiman}{bme_technion}
	\end{icmlauthorlist}
		
	\icmlaffiliation{dds_technion}{
		Faculty of Data and Decision Sciences, Technion --- Israel Institute of Technology,
		Haifa, Israel
	}
	\icmlaffiliation{cs_technion}{
		Faculty of Computer Science, Technion --- Israel Institute of Technology,
		Haifa, Israel
	}
	\icmlaffiliation{bme_technion}{
		Faculty of Biomedical Engineering, Technion --- Israel Institute of Technology,
		Haifa, Israel
	}
		
	\icmlcorrespondingauthor{Zvi N. Badash}{zvi.badash@campus.technion.ac.il}
		
	\icmlkeywords{Uncertainty Estimation, Information Theory, LLMs, Large Language Models}
		
	\vskip 0.3in
]



\printAffiliationsAndNotice{}  

\begin{abstract}
	Large language models (LLMs) are often confidently wrong, making reliable uncertainty estimation (UE) essential. Output-based heuristics are cheap but brittle, while probing internal representations is effective yet high-dimensional and hard to transfer.
	We propose a compact, per-instance UE method that scores cross-layer agreement patterns in internal representations using a single forward pass.
	Across three models, our method matches probing in-distribution, with mean diagonal differences of at most $-1.8$ AUPRC percentage points and $+4.9$ Brier score points. Under cross-dataset transfer, it consistently outperforms probing, achieving off-diagonal gains up to $+2.86$ AUPRC and $+21.02$ Brier points. Under 4-bit weight-only quantization, it remains robust, improving over probing by $+1.94$ AUPRC points and $+5.33$ Brier points on average.
	Beyond performance, examining specific layer--layer interactions reveals differences in how disparate models encode uncertainty. Altogether, our UE method offers a lightweight, compact means to capture transferable uncertainty in LLMs.
\end{abstract}

\section{Introduction}\label{sec:intro}
Large language models (LLMs) are increasingly deployed in domains where erroneous answers can have tangible costs. Yet, despite striking task performance, LLMs often produce confident, grammatically fluent, but \emph{incorrect} outputs (``hallucinations''). This miscalibration, in which high confidence is assigned to wrong predictions, undermines reliability in knowledge-intensive or safety-critical settings and motivates principled \emph{uncertainty estimation} (UE) at inference time.

A natural family of UE methods scores the model's \emph{outputs}. Token-probability heuristics (e.g., entropy or margin) are simple and fast, but can be brittle, conflating lexical surface forms with semantic confidence and failing under distribution shift. Bayesian surrogates (e.g., MC Dropout \cite{GalGhahramani2016} or deep ensembles \cite{Lakshminarayanan2017}) are more expressive but computationally expensive at scale, often making this approach computationally prohibitive. In parallel, \emph{probing} work has shown that LLM internals encode signals correlated with correctness; however, these probes often rely on high-dimensional hidden vectors, tend to be task-specific, and can be difficult to interpret or generalize \citep{AzariaMitchell2023, orgad2025llms, bar-shalom2025beyond, bar2025learning}.

We take a different route: \emph{structure} the internal signal before learning from it. Concretely, we view each layer's post-MLP activation as a probability distribution (via temperature-scaled softmax over the hidden dimension) and compute a pairwise, directed KL divergence between layers at task-relevant tokens. The resulting $L{\times}L$ \emph{signature} compactly captures cross-layer (dis)agreement. A small gradient-boosted tree (GBDT) model, trained on these maps, predicts whether the model's answer is correct; we use its output score for uncertainty. Conceptually, our approach lies \emph{between} classic probing and the Information Bottleneck (IB) perspective \citep{Tishby1999IB, Alemi2017VIB, Achille2018IB}: it uses local, inference-time, representation-level structure grounded in information theory, without attempting to estimate global mutual informations (e.g., $I(X;T)$) that typically require estimating the joint distribution of multi-dimensional vectors and are impractical online.

\begin{figure*}[t]
	\centering
		
	\begin{subfigure}[t]{0.45\linewidth}
		\centering
		\includegraphics[width=\linewidth]{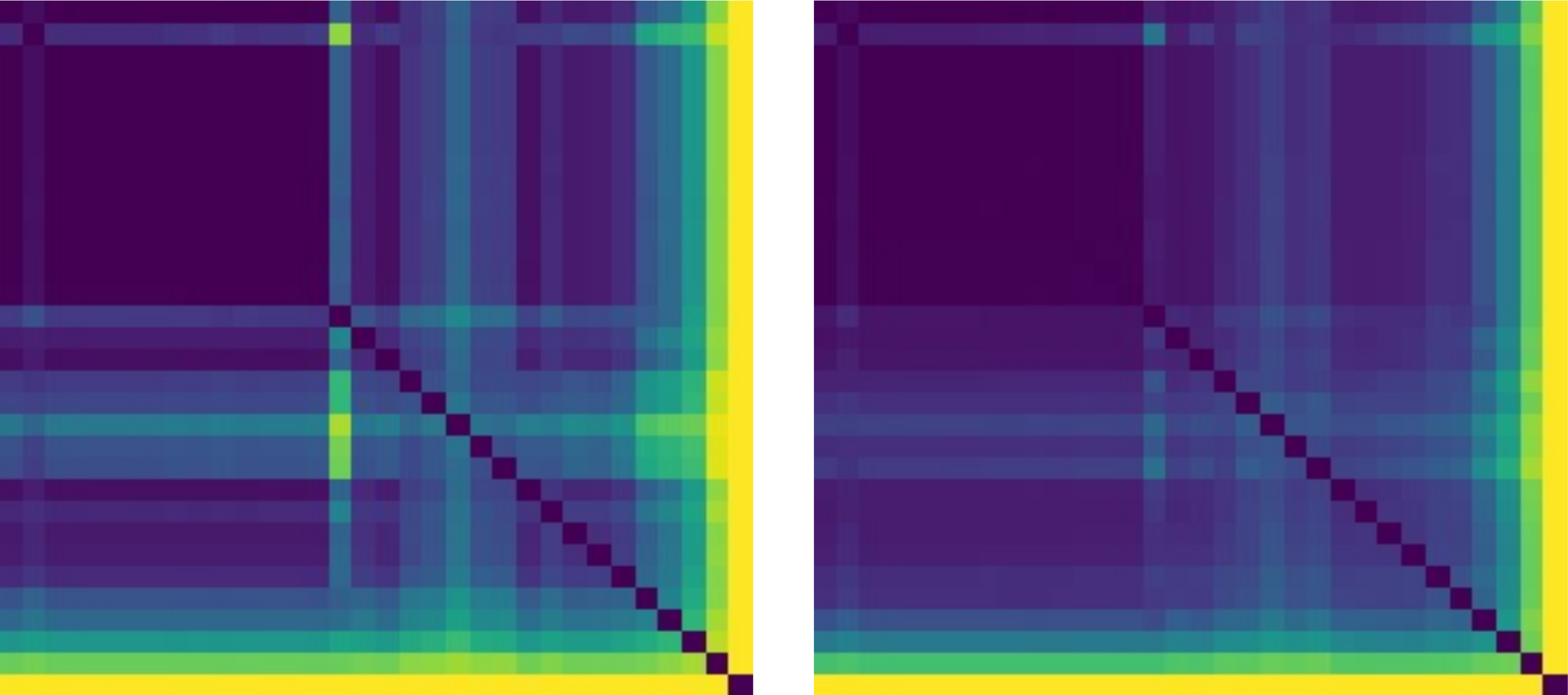}
		\caption{Incorrect predictions (random samples)}
		\label{fig:mmlu-mistral-incorrect}
	\end{subfigure}\hfill
	\begin{subfigure}[t]{0.45\linewidth}
		\centering
		\includegraphics[width=\linewidth]{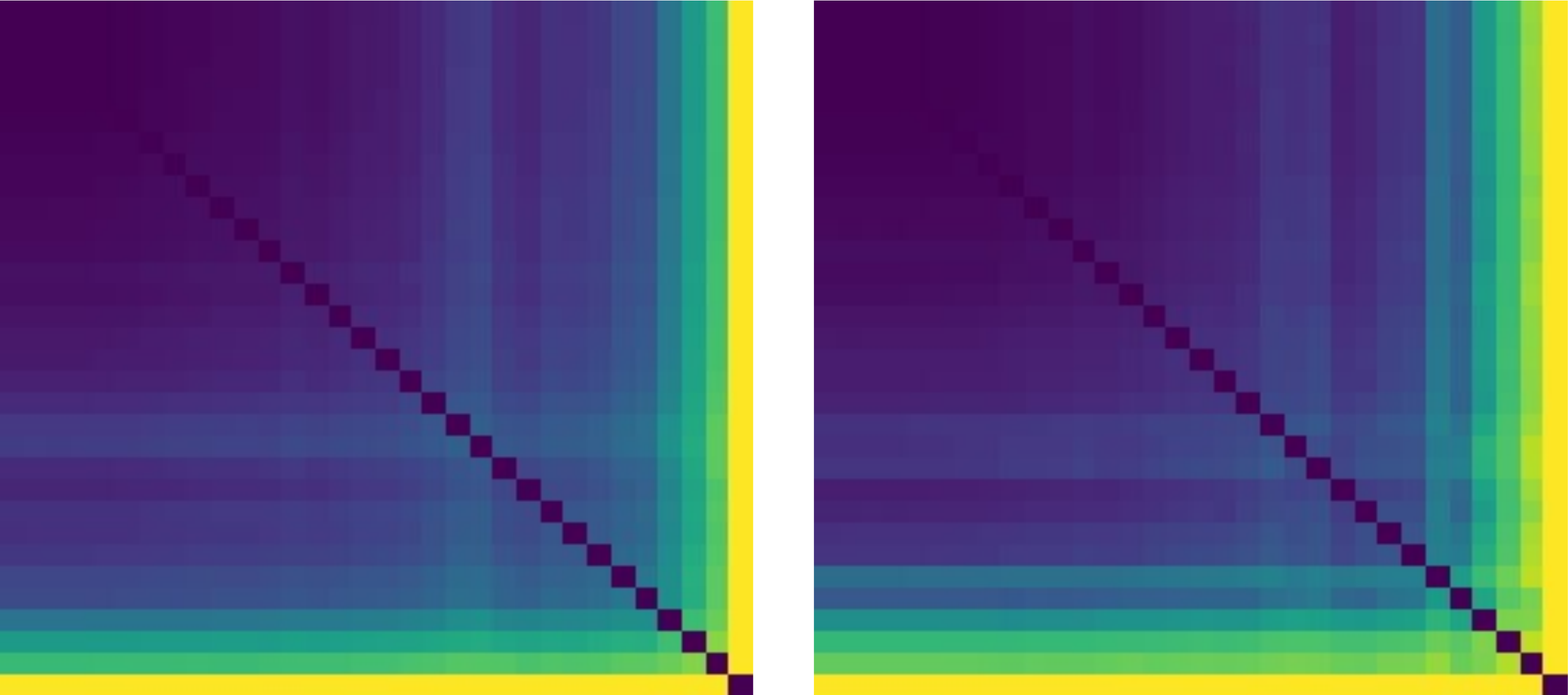}
		\caption{Correct predictions (random samples)}
		\label{fig:mmlu-mistral-correct}
	\end{subfigure}
		
	\caption{
		KL \emph{signature maps} on \textsc{MMLU} for \textbf{Mistral-7B-Instruct-v0.3}.
		Each heatmap shows an $L{\times}L$ matrix of directed divergences computed at task-relevant tokens; warmer colors indicate larger $D_{\mathrm{KL}}$.
		Panels (a) and (b) show random examples from incorrect and correct predictions, respectively, highlighting differences in cross-layer agreement patterns.
	}
	\label{fig:mmlu-mistral-signatures}
\end{figure*}

\paragraph{Contributions.}
\begin{enumerate}
	\item \textbf{Layer-wise, information-theoretic signatures.} KL-based, directed layer-to-layer signature maps as a structured representation of neuronal activations at task-relevant tokens. We provide an example of such signatures in Figure~\ref{fig:mmlu-mistral-signatures}.
	\item \textbf{Compact estimator.} A lightweight GBDT trained on signature maps to produce a per-instance score (no architectural changes or multiple forward passes).
	\item \textbf{Evaluation suite.} Experiments on the datasets used by \citet{orgad2025llms} across three models (Llama-3.1-8B, Qwen3-14B-Instruct, and Mistral-7B-Instruct-v0.3)---\textsc{TriviaQA}, \textsc{HotpotQA} (without context), \textsc{Movies}, \textsc{WinoGrande}, \textsc{WinoBias}, \textsc{IMDB}, \textsc{Math} (answerable subset)---and additionally on \textsc{MMLU}.
\end{enumerate}

\noindent We argue that signature maps offer a structured perspective on how agreement patterns evolve across depth; a comprehensive interpretability study is left for future work.

\section{Background}\label{sec:background}

\paragraph{Why hallucinations persist.}
Recent analysis argues that current training and \emph{evaluation} setups often reward confident guessing over calibrated admission of uncertainty, especially on tasks that grade one-shot answers without credit for abstention or evidence \citep{OpenAIWhyHallucinate2025}. Under such incentives, an LLM that is uncertain may still prefer a fluent guess to ``I don't know.'' This reframes hallucinations not as rare failures but as an expected consequence of objectives--making \emph{inference-time} uncertainty estimation (UE) central to safe deployment.

\subsection{Uncertainty in LLMs}\label{sec:bg-ue}
Let $p_\theta(y\mid x)$ denote a model's predictive distribution. A basic confidence proxy is \emph{predictive entropy},
\[
	H\!\left[p_\theta(\cdot\mid x)\right] \;=\; - \sum_{y} p_\theta(y\mid x)\,\log p_\theta(y\mid x),
\]
and, in classification settings, the \emph{margin} between the top two classes,
$m(x)=p_\theta(\hat y\mid x)-p_\theta(y_{(2)}\mid x)$.
Such output-based heuristics are computationally cheap and widely used, but they are sensitive to paraphrasing, prompt format, and spurious surface cues. More semantic or meaning-aware uncertainty measures improve robustness at the cost of additional generations or model calls \citep{Zablotskaia2023, Farquhar2024}, and all such approaches struggle with high-confidence hallucinations \citep{simhi2025trust}. More recent work has sought to strengthen output-based uncertainty estimation by exploiting richer observable signals. In particular, Bar-Shalom et al.\citep{bar2025learning} propose modeling the full sequence of next-token probability distributions---rather than only realized token probabilities---as an \emph{LLM Output Signature} (LOS). 

Despite these advances, output-only methods---whether heuristic or learned---are fundamentally limited to information exposed at the surface level and may miss internal failure signals that do not manifest clearly in the output distribution.

\subsection{Probing internal representations}\label{sec:bg-probing}
\emph{Probing} methods investigate what information is encoded in intermediate representations by training an auxiliary classifier $g_\phi$ on hidden states $h_\ell^{(t)}$ to predict a property $z$ (e.g., syntactic features, semantic relations, or correctness):
\[
	\min_{\phi}\; \mathbb{E}_{(x,z)}\big[\mathcal{L}\big(g_\phi(h_\ell^{(t)}(x)),\, z\big)\big].
\]
Foundational work used probing to map linguistic and semantic information across layers and tokens, alongside controls to disentangle representational content from probe capacity \citep[][inter alia]{hupkes2018visualisation,BelinkovGlass2019, Belinkov2022}. More recently, probes have been applied to reliability-related targets, including truthfulness and correctness estimation \citep{AzariaMitchell2023,kadavath2022language,liu2023cognitive,gottesman2024estimating, steyvers2025large}.

While probing offers access to internal signals unavailable to output-only methods, practical challenges remain. Hidden states are high-dimensional and difficult to interpret, probe performance can be sensitive to the choice of layer and token, and transfer across datasets or tasks is often limited \citep{orgad2025llms}. These limitations motivate approaches that retain probing’s per-instance access to internal evidence while imposing additional structure to improve interpretability, robustness, and cross-task generalization.

\subsection{Methods considering layer interactions}
Several prior works exploit interactions between intermediate layer representations to address reliability or reasoning failures in Transformer models, including BLOOD \citep{BLOODjelenic2024out}, Seq-VCR \citep{SEQVCRarefinseq}, and ACT-ViT \citep{bar-shalom2025beyond}. These methods leverage structural regularities across layers but rely on task-specific assumptions about the ``natural'' behavior of representations at inference time.

BLOOD (Between-Layer Out-of-Distribution Detection) \citep{BLOODjelenic2024out} is a post-hoc OOD detection method that requires neither training data nor model retraining. It exploits the empirical observation that between-layer transformations are smoother for in-distribution inputs than for OOD inputs, quantifying sharpness via the squared Frobenius norm of the Jacobian,
$\phi_\ell(\boldsymbol{x}) = \| \boldsymbol{J}_\ell(\boldsymbol{h}_\ell) \|_F^2$ (or an unbiased estimator), and detecting OOD inputs through deviations in this smoothness profile. The method thus assumes that smooth inter-layer mappings characterize in-distribution behavior.

Seq-VCR (Sequential Variance--Covariance Regularization) \citep{SEQVCRarefinseq} addresses representation collapse in decoder-only Transformers, which degrades multi-step reasoning, particularly in arithmetic tasks. It introduces a training-time regularizer that increases the entropy of intermediate representations by encouraging diversity in their variance--covariance structure across layers. Combined with dummy pause tokens as a lightweight alternative to chain-of-thought supervision, Seq-VCR improves reasoning by preventing low-rank or redundant intermediate states.

ACT-ViT \citep{bar-shalom2025beyond} targets hallucination detection by modeling joint structure across layers and tokens. Rather than probing isolated layer--token pairs, it treats activation tensors as structured objects and applies a Vision Transformer-style architecture to capture global activation patterns, enabling cross-model training and transfer under the assumption that hallucination signals manifest as stable spatial patterns in activation space.

\subsection{Information Bottleneck (IB) perspectives}\label{sec:bg-ib}
Whereas the preceding methods extract \emph{instance-level} signals from interactions between intermediate representations, the Information Bottleneck (IB) provides an information-theoretic framework for analyzing how task-relevant information \emph{flows} through a network, emphasizing representation dynamics rather than static views of individual layers. Specifically, the IB frames representation learning as a trade-off between preserving task-relevant information and compressing inputs:
\[
	\max_{p(t\mid x)}\;\; I(T;Y)\;-\;\beta\, I(X;T),
\]
with $T$ a stochastic representation of $X$ \citep{Tishby1999IB}. Information-plane analyses in deep nets (tracking $(I(X;T),I(T;Y))$ during training) suggest phases of fitting and compression, sparking debates on mechanisms and estimators \citep{ShwartzZivTishby2017, Saxe2018, Achille2018IB}. Crucially, these information measures are defined at the \emph{distribution} level and require aggregation over many samples, making them impractical for \emph{single-instance} decisions at inference time. This creates a gap between IB's global analysis and the local decisions needed to modulate model behavior on a specific input.

\begin{figure*}[t!]
	\centering
	\includegraphics[width=.78\linewidth]{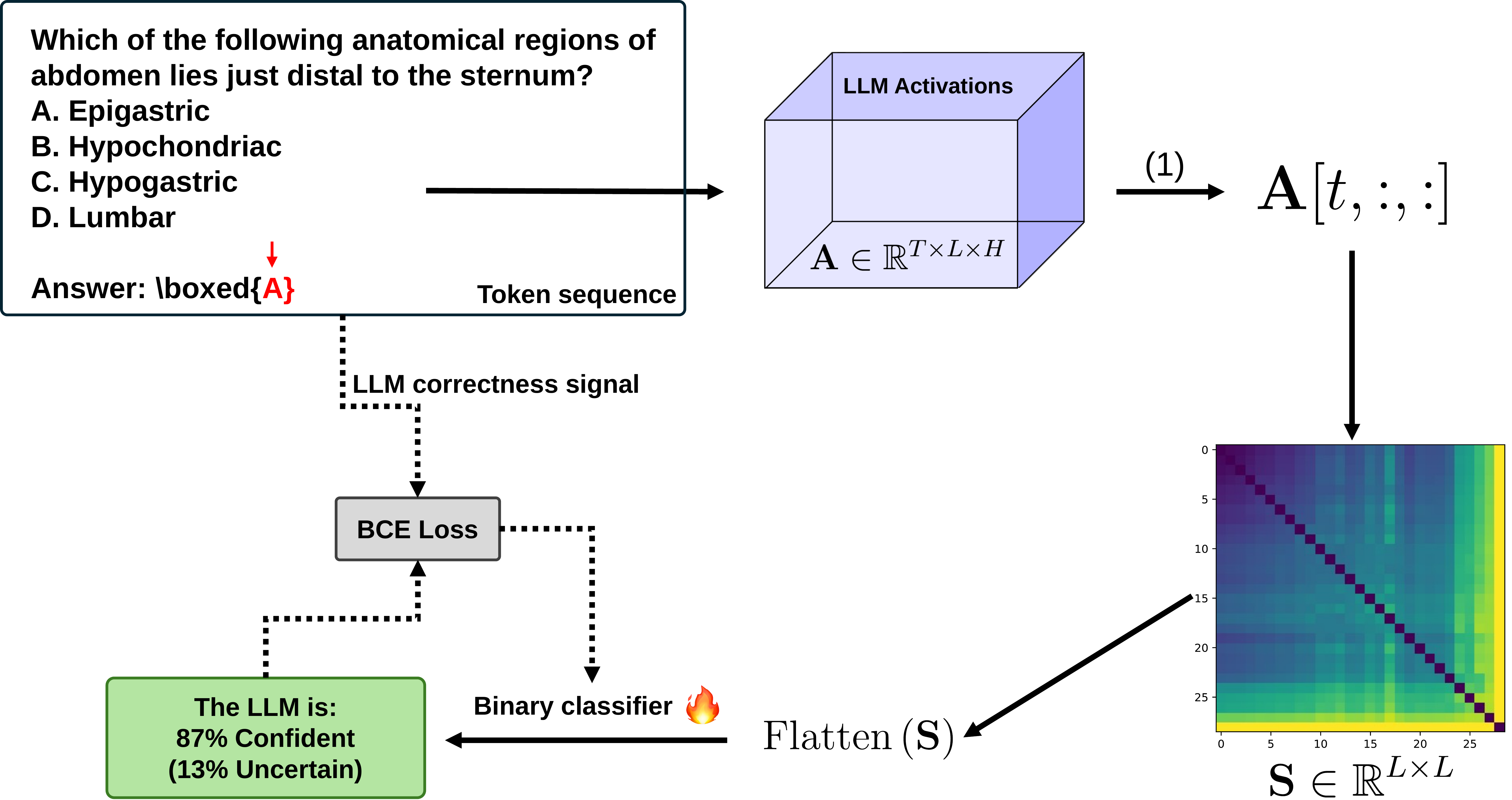}
	\caption{\textbf{Pipeline overview.}
		Post-MLP activations are normalized via softmax to distributions $\boldsymbol{p}_\ell^{(t)}$; pairwise directed KLs produce an $L{\times}L$ signature map per token; after optional contrast correction and flattening, a LightGBM classifier outputs a per-instance score. Stage (1) signifies sampling tokens to perform analysis on. Strategies include looking at exact answer tokens \citep{orgad2025llms}, looking at the last token in the sequence, or aggregating activations over several tokens.}
	\label{fig:method}
\end{figure*}

\subsection{Positioning}
Existing approaches that exploit interactions between intermediate layers (i.e. BLOOD and Seq-VCR), encode task-agnostic assumptions about the expected behavior of representations (e.g., smoothness and entropy). While effective in specific regimes, these fixed priors may not generalize across tasks or inference conditions.

Our approach departs from this paradigm by avoiding assumptions about the dynamics of representation. Instead, it infers task-dependent information flow \emph{at inference time} by modeling directed, pairwise interactions between \emph{all} layers, enabling the capture of long-range dependencies across depth rather than relying on local or sequential relationships.

A complementary line of work modifies \emph{decoding} rather than estimating reliability post hoc, including DoLa \citep{chuang2023dola} and uncertainty-aware decoding methods based on minimum Bayes risk or selective generation \citep{cheng-vlachos-2023-faster, daheim2025uncertaintyaware}. These approaches alter generation dynamics but do not produce per-instance uncertainty estimates.

By contrast, we perform post-hoc reliability estimation by \emph{reading} structured internal evidence. Each instance is summarized via directed, pairwise divergences between layer representations, yielding a compact $L{\times}L$ signature with $L^2 \ll d_{\text{hidden}}$ in modern LLMs. This representation emphasizes cross-layer agreement and disagreement rather than absolute activations, and we hypothesize that it has lower \emph{effective} information dimension while preserving task-relevant correctness signals.

Overall, our method instantiates an uncertainty estimation framework that operates \emph{per instance} on internal evidence, imposes an interpretable information-theoretic structure, and avoids dataset-level mutual-information estimators and restrictive priors on layer behavior. The resulting calibrated correctness probabilities support abstention \citep{Chow1970}, triage, calibration, and can serve as a principled trigger for uncertainty-aware decoding methods such as DoLa.

\section{Methodology}\label{sec:method}
Our approach estimates uncertainty from intrinsic signals by exploiting statistical relationships between intermediate layer representations. The pipeline comprises: (i) transforming hidden activations into probability distributions, (ii) constructing layer-wise signature maps via pairwise KL divergences, and (iii) training a lightweight classifier to predict correctness from these maps. See Figure~\ref{fig:method}.

\subsection{Problem Framing and Notation}\label{sec:prob-def-notation}
Let $f_{\theta}$ be an LLM that, given an input $x$, produces an output $\hat{y}(x)$ and internal activations $\{\boldsymbol{h}^{(t)}_{\ell}(x)\in\mathbb{R}^{d_{\mathrm{model}}}\}_{\ell=1}^{L}$ for token positions $t\in\{1,\dots,T\}$. We evaluate on $\mathcal{D}=\{(x_i,c_i)\}_{i=1}^{n}$, where $c_i\in\{0,1\}$ indicates whether $\hat{y}(x_i)$ is \emph{correct} under the dataset-specific criteria in \S\ref{app:datasets}. A deterministic selector $\mathcal{I}(x)\subseteq\{1,\dots,T\}$ identifies task-relevant positions (e.g., last token for classification or exact answer tokens for QA, following \cite{orgad2025llms}).

We denote by $\Phi$ the feature map that converts the selected internal signals into a compact representation $\boldsymbol{z}(x)\in\mathbb{R}^{m}$ (via the layer--layer signature pipeline detailed below). A small classifier $g_{\psi}$ then returns a \emph{confidence}
\[
	q(x)\;=\;P_{\psi}\!\big(\text{correct}\mid \boldsymbol{z}(x)\big)\in[0,1],
\]
and we report the corresponding \emph{uncertainty} as
\[
	u(x)\;=\;1-q(x).
\]
We assess two complementary aspects. First, we ask whether $u(\cdot)$ can \emph{separate} mistakes from correct predictions without choosing a threshold; we summarize this error-detection view with \textbf{AUPRC}, which is robust to class imbalance (\S\ref{sec:setup-metrics}). Second, we evaluate whether $q(\cdot)$ behaves like a probability. We quantify probabilistic quality with the \textbf{Brier score} defined here as \emph{$1$ minus the standard Brier loss} (i.e., $1-\mathrm{MSE}$ between predicted correctness probabilities and binary ground truth), so that higher is better.

\subsection{Layer-wise Activation Distributions}\label{sec:method-dists}
Consider an LLM with $L$ transformer layers, each producing hidden states of dimension $d_{\text{model}}$. For an input sequence $\{x_1, \ldots, x_T\}$, we extract the activations after each transformer block $h_\ell^{(t)} \in \mathbb{R}^{d_{\text{model}}}$ for some token $t$ at every layer $\ell \in \{1,\ldots,L\}$. We apply a temperature-scaled softmax to each vector:

\vspace{-1.3em}
\[
	\boldsymbol{p}_\ell^{(t)} = \operatorname{Softmax}\!\left(\frac{\boldsymbol{h}_\ell^{(t)}}{\tau}\right), \qquad
	\boldsymbol{p}_\ell^{(t)} \in \Delta^{d_{\text{model}}-1},
\]
\vspace{-.9em}

where $\tau>0$ is a global temperature parameter and $\Delta^{d_{\text{model}}-1}$ is the simplex.
This normalization makes the vectors amenable for probabilistic distance measures defined below, although other normalizations are possible in principle. 

\subsection{Signature Maps via KL Divergence}\label{sec:method-sig}
For token $t$, we define the directed divergence matrix $\textbf{S}^{(t)} \in \mathbb{R}^{L{\times}L}$ as:
\[
	S_{ij}^{(t)} = D_{\mathrm{KL}}\!\big(\boldsymbol{p}_i^{(t)} \,\|\, \boldsymbol{p}_j^{(t)}\big), \qquad i,j \in \{1,\ldots,L\}.
\]
The ``distance'' used in this formulation is $D_\text{KL}$, but can be easily extended to other quantities, such as the symmetric Jensen-Shannon divergence, $$
D_{\text{JS}} := \frac{1}{2}D_\text{KL}\left(p\|m\right) + \frac{1}{2}D_\text{KL}\left(q\|m\right), \text{ where } m:=\frac{p+q}{2},$$
for example. Our implementation supports both versions. A quantitative comparison can be found in \S\ref{sec:jsd-vs-kl}. 

We focus on task-relevant tokens following \citet{orgad2025llms}: for QA, either the final generated token or exact answer tokens (obtained via an auxiliary procedure); for classification, the predicted class token. Optionally, after constructing the layer--layer divergence matrix, we apply an element-wise contrast transformation to improve the uniformity and dynamic range of the resulting maps; the strength of this transformation is controlled by a tunable hyperparameter $\alpha$.
\[
	\textbf{S}^{\prime (t)} = 1 - \exp\!\big(-\alpha \, \textbf{S}^{(t)}\big), \qquad \alpha>0,
\]
We then flatten to features $\boldsymbol{z}^{(t)}=\operatorname{Flatten}\!\left(S^{\prime (t)}\right)$.

\subsection{Uncertainty Estimation via Gradient Boosting}\label{sec:method-gbdt}
We train a LightGBM classifier \citep{Ke2017LightGBM} per task on $z^{(t)}$ to predict correctness. Given $z^{(t)}$, we use
\[
	\hat{u}(\boldsymbol{z}^{(t)}) \;=\; 1 - P\!\left(\text{correct}\mid \boldsymbol{z}^{(t)}\right)
\]
as the uncertainty score. We report raw probabilities and do not apply post-hoc calibration; standard post-hoc methods such as Platt scaling \citep{Platt1999} or isotonic regression \citep{ZadroznyElkan2001,ZadroznyElkan2002} are compatible but left for future work.

\section{Experimental Setup}\label{sec:setup}
\subsection{Models \& datasets.}
We evaluate three models: Llama-3.1-8B (base, non-instruct), Qwen3-14B-Instruct, and Mistral-7B-Instruct-v0.3, across the benchmark suite used by \citet{orgad2025llms}: 
\textsc~{TriviaQA} \citep{TriviaQA2017}, \textsc~{HotpotQA} with and without context \citep{HotpotQA2018}, \textsc~{Movies} (as defined in \citealp{orgad2025llms}), \textsc~{WinoGrande} \citep{WinoGrande2020}, \textsc{WinoBias} \citep{WinoBias2018}, \textsc~{MNLI} \citep{MNLI2018}, \textsc{IMDB} \citep{IMDB2011}, \textsc~{Math} (answerable subset; we follow the \textsc~{GSM8K} protocol \citealp{GSM8K2021}), \textsc~{Natural Questions} with context \citep{NQ2019}, and \textsc~{MMLU} \citep{MMLU2021}. Details are available in \S\ref{app:data-models}.  

\subsection{Baselines}\label{sec:setup-baselines}
We compare our method primarily to linear-\textbf{probing}-based uncertainty estimation following the protocol of \citet{orgad2025llms}. This choice is deliberate. Probing is the closest conceptual and operational baseline to our approach: both methods extract signals from internal representations at inference time, require supervision, operate per instance, and incur comparable computational cost (a single forward pass plus a lightweight classifier).

While our primary point of comparison is probing, a full evaluation of more expressive approaches that operate on hidden representations across all token positions---such as \citet{bar-shalom2025beyond,bar2025learning}---is beyond the scope of this work. Nevertheless, we include a direct comparison with LOS-NET (Table~\ref{tab:signatures_vs_losnet}) and contrast representational efficiency with LOS-NET \citep{bar2025learning} and ACT-ViT \citep{bar-shalom2025beyond}.

Other families of uncertainty estimation methods are not directly comparable at the scale considered here. Output-based heuristics (e.g., entropy or margin) are substantially cheaper but are known to be brittle under distribution shift and do not use internal signals. Bayesian and sampling-based methods (e.g., MC Dropout or ensembles) require multiple forward passes and are computationally prohibitive for large models across the multi-dataset, multi-model setting we study. Generation-based semantic uncertainty methods introduce additional decoding steps and prompt sensitivity, making them difficult to control and fairly compare.

By focusing on probing, we isolate the effect of \emph{representation structure} rather than model access, compute budget, or decoding strategy. This ensures a fair comparison under identical inference-time constraints, highlighting whether structured layer--layer signatures provide an advantage over raw hidden-state features when scale and compute are held fixed.

\subsection{Metrics}\label{sec:setup-metrics}
We report two metrics. For threshold-free error detection we use \textbf{AUPRC}, which is suitable under label imbalance and reflects ranking quality of errors versus correct predictions. For probabilistic quality we report the \textbf{Brier score} defined as \emph{$1$ - Brier loss} (i.e., $1-\mathrm{MSE}$ of predicted correctness probabilities); this makes the score comparable in direction to AUPRC (higher is better).

\section{Experiments and Results}\label{sec:experiments}
\subsection{E1: In-distribution performance}\label{sec:exp-e1}
We first examine in-distribution performance, training and evaluating on the same dataset for each model. Table~\ref{tab:e2} (diagonal entries) summarizes the comparison between our \emph{signature} estimator and \emph{probing} across all three architectures.

Overall, the two methods exhibit near parity in the in-distribution setting. While probing achieves consistently higher \textbf{AUPRC} in distribution, our method yields consistently superior \textbf{Brier scores}, indicating better calibrated correctness probabilities despite slightly weaker ranking of errors. This trade-off is stable across models and datasets, suggesting that the structured layer--layer representation preserves probabilistic information about correctness while sacrificing a small amount of discriminative power when train and test distributions coincide.

Taken together, these results show that compression into $L\times L$ signature maps does not materially degrade in-distribution uncertainty estimation, and can even improve probabilistic calibration relative to probing.

\subsection{E2: Cross-task generalization}\label{sec:exp-e2}
We now test \emph{cross-task} transfer by training the estimator on task $X$ and evaluating it on a \emph{different} task $Y$, for all ordered pairs with aligned token selection and evaluation.

\begin{figure*}[t!]
	\centering
	\begin{subfigure}{0.38\textwidth}
		\centering
		\includegraphics[width=\linewidth]{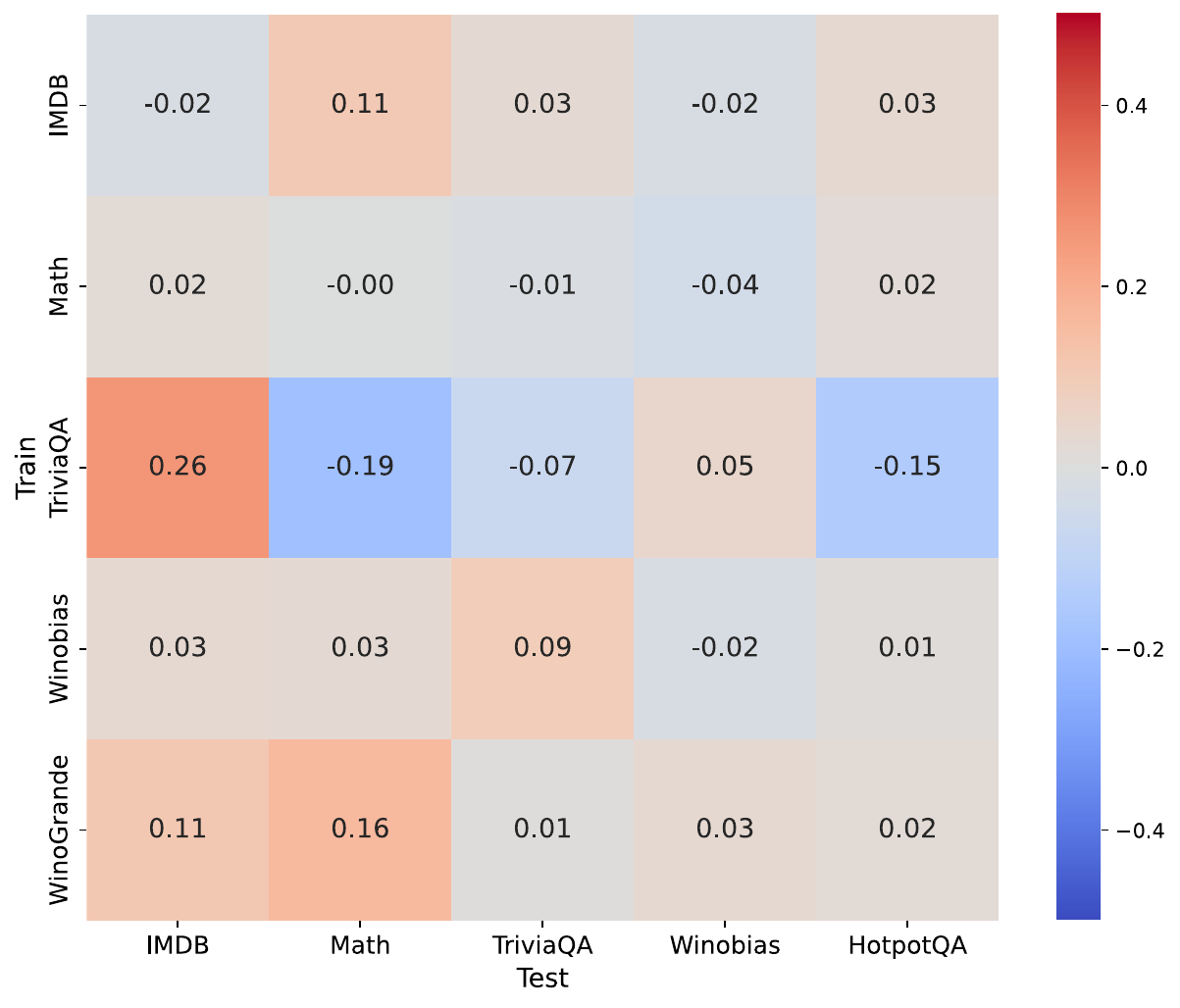}
		\caption{\centering AUPRC difference values  (Llama-3.1-8B).\\Diagonal mean difference: $-0.0180$.\\Off-diagonal mean difference: $0.0286$.}
		\label{fig:e2_pr_llama}
	\end{subfigure}\hspace{3cm}
	\begin{subfigure}{0.38\textwidth}
		\centering
		\includegraphics[width=\linewidth]{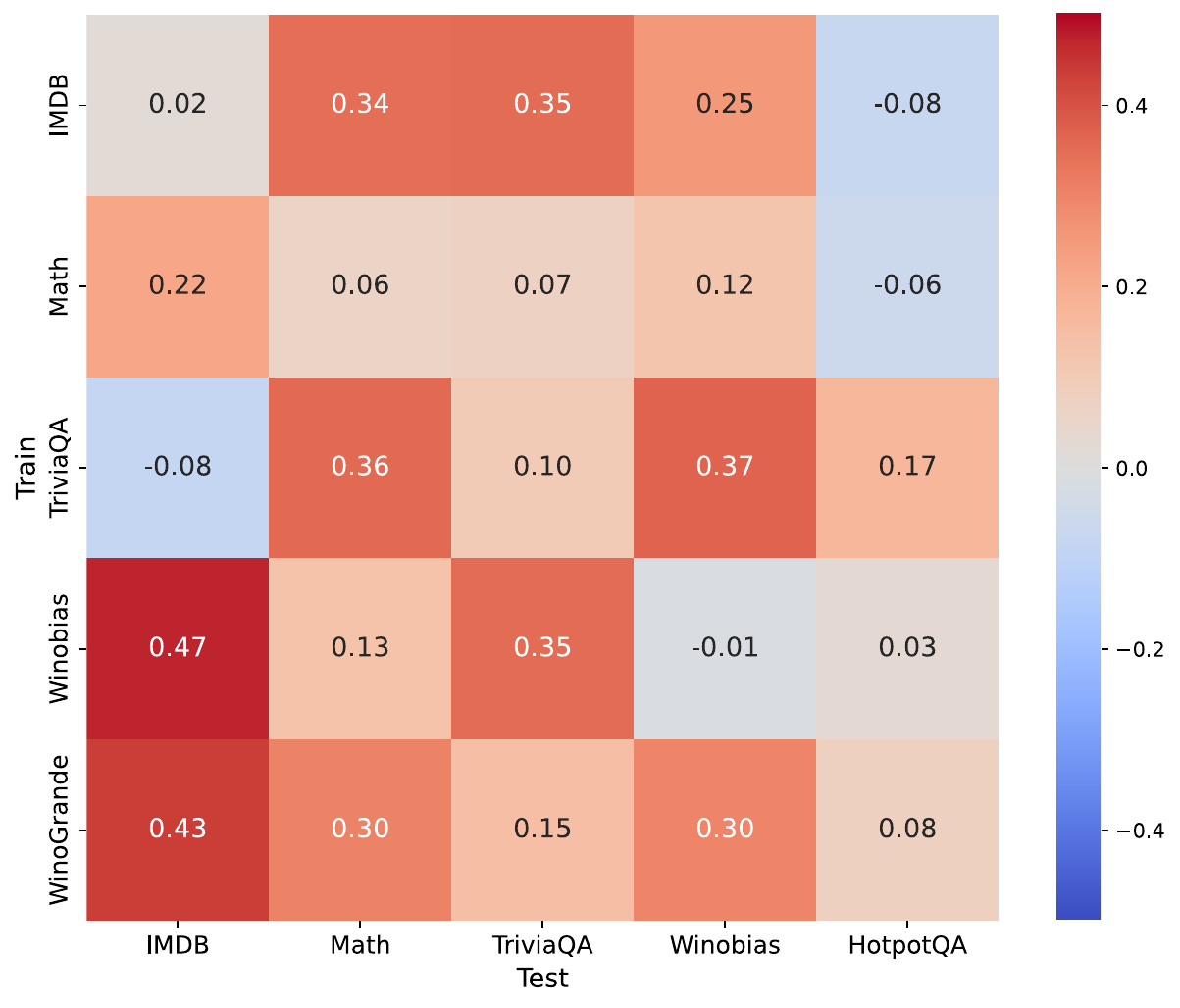}
		\caption{\centering Brier score difference values (Llama-3.1-8B).\\Diagonal mean difference: $0.0488$.\\Off-diagonal mean difference: $0.2102$.}
		\label{fig:e2_brier_llama}
	\end{subfigure}
		
	\begin{subfigure}{0.38\textwidth}
		\centering
		\includegraphics[width=\linewidth]{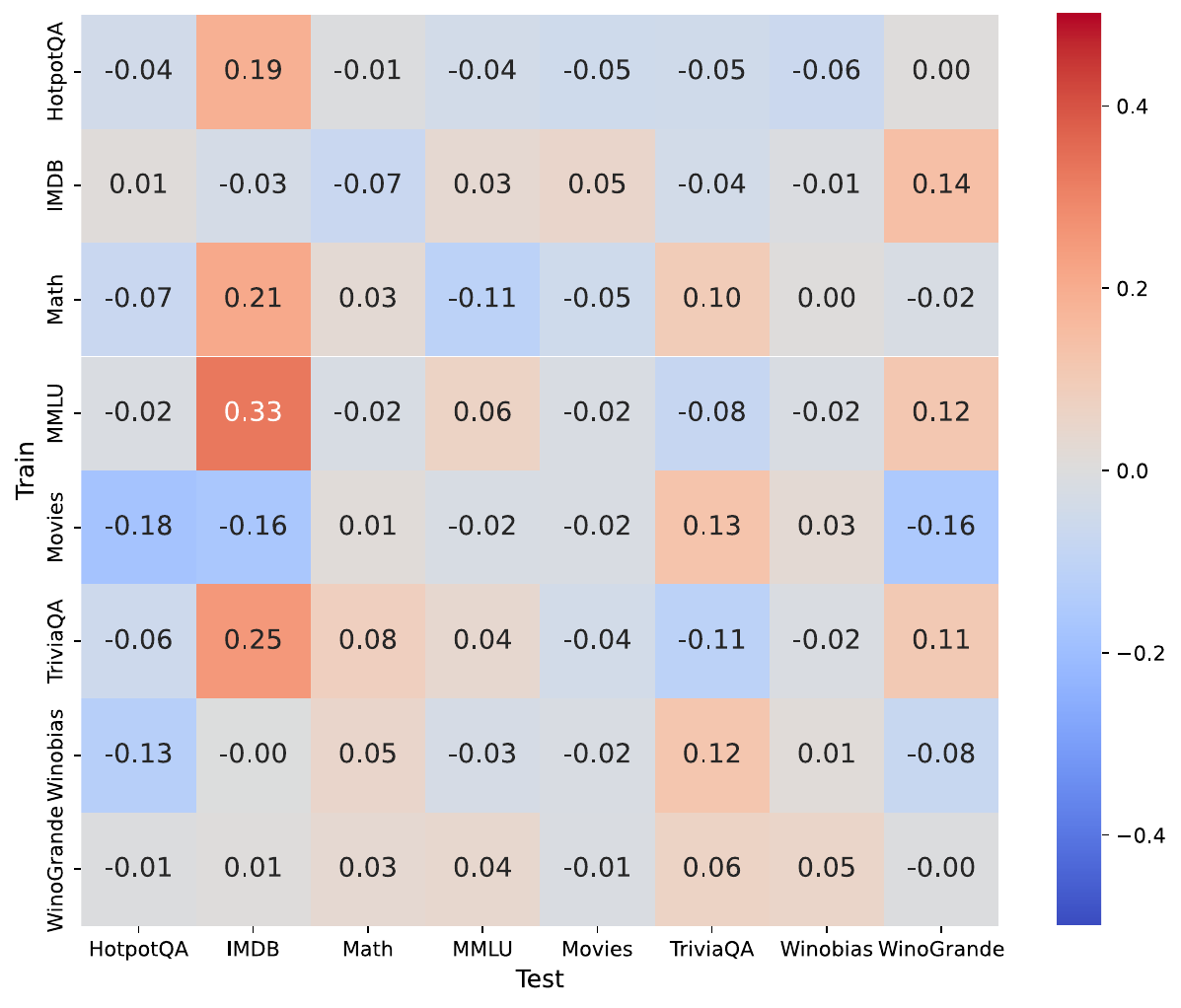}
		\caption{\centering AUPRC difference values (Qwen3-14B).\\Diagonal mean difference: $-0.0137$.\\Off-diagonal mean difference: $0.0095$.}
		\label{fig:e2_prauc_qwen}
	\end{subfigure}\hspace{3cm}
	\begin{subfigure}{0.38\textwidth}
		\centering
		\includegraphics[width=\linewidth]{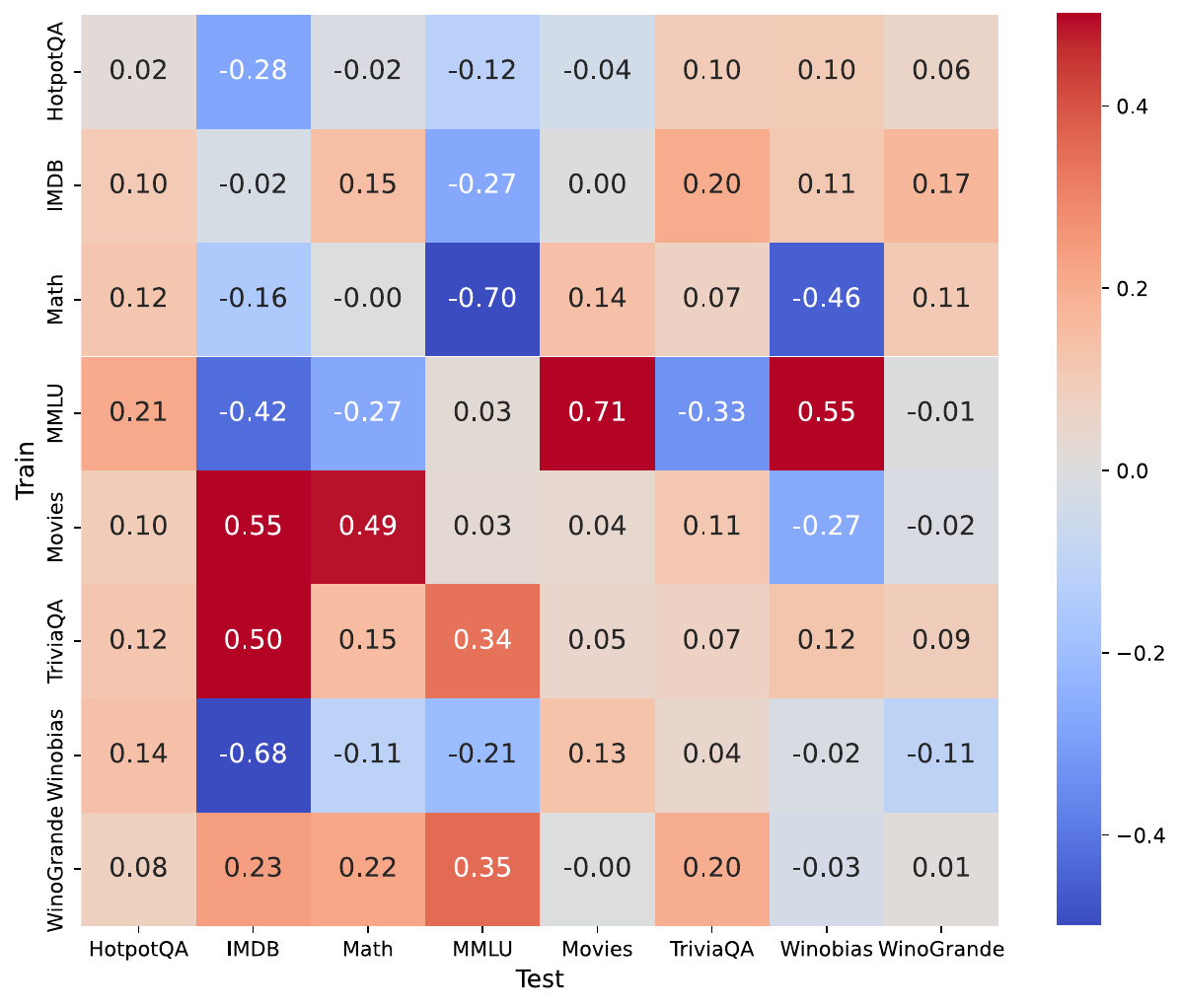}
		\caption{\centering Brier score difference values (Qwen3-14B).\\Diagonal mean difference: $0.0152$.\\Off-diagonal mean difference: $0.0435$.}
		\label{fig:e2_brier_qwen}
	\end{subfigure}
		
	\begin{subfigure}{0.38\textwidth}
		\centering
		\includegraphics[width=\linewidth]{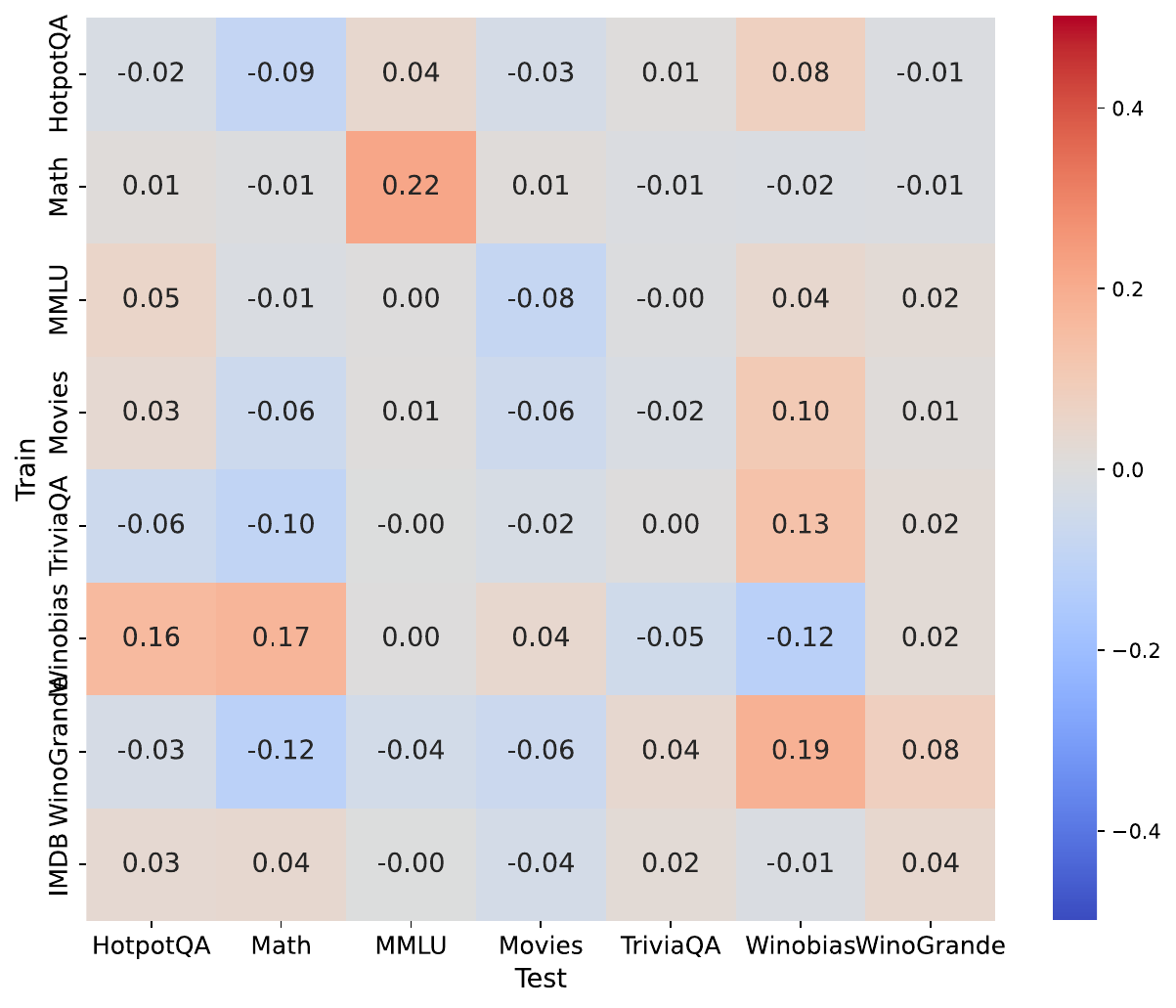}
		\caption{\centering AUPRC difference values (Mistral-7B-v0.3-Instruct).\\Diagonal mean difference: $-0.0172$.\\Off-diagonal mean difference: $0.0135$.}
		\label{fig:e2_prauc_mistral}
	\end{subfigure}\hspace{3cm}
	\begin{subfigure}{0.38\textwidth}
		\centering
		\includegraphics[width=\linewidth]{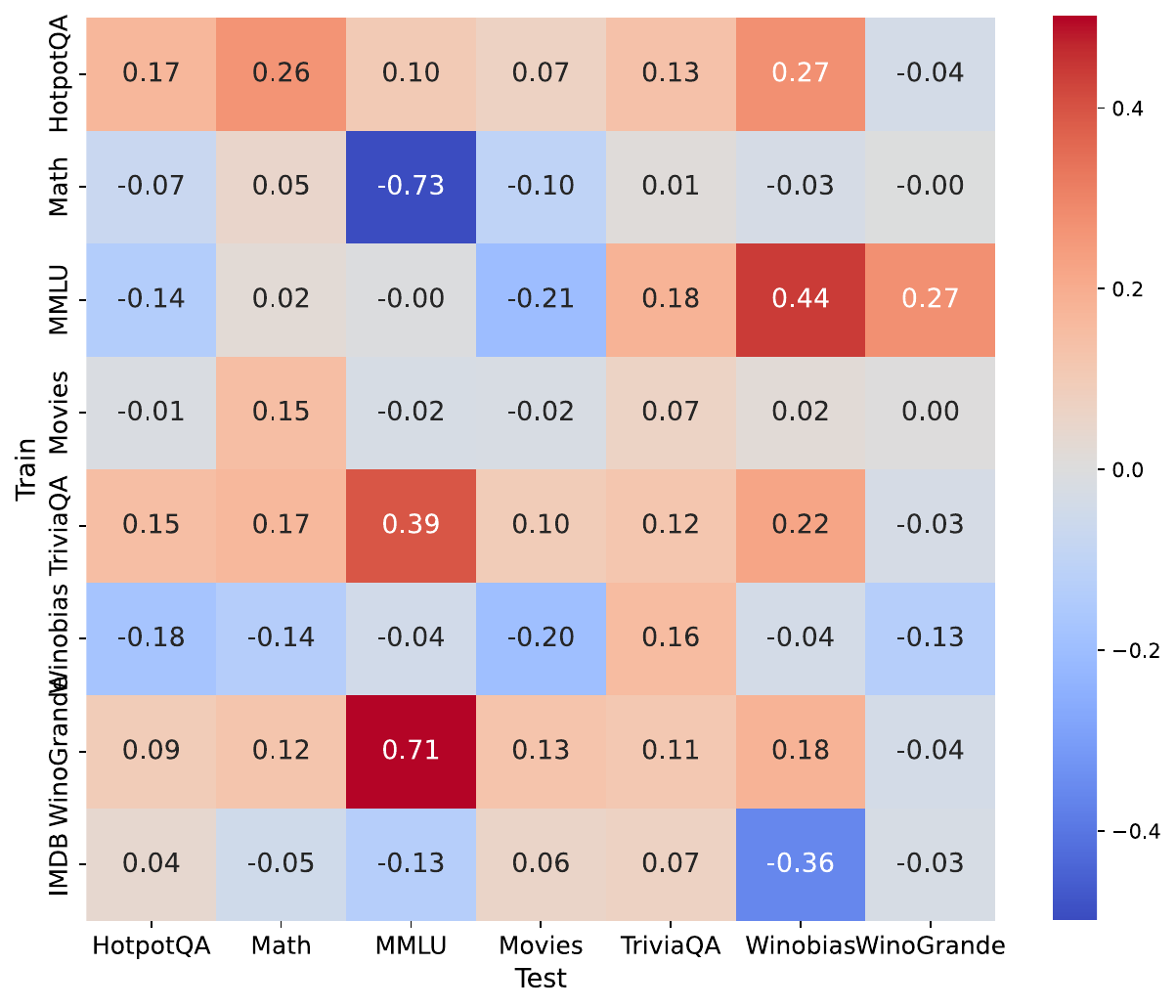}
		\caption{\centering Brier score difference values (Mistral-7B-v0.3-Instruct).\\Diagonal mean difference: $0.0338$.\\Off-diagonal mean difference: $0.0428$.}
		\label{fig:e2_brier_mistral}
	\end{subfigure}
	\caption{
		Cross-\emph{task} generalization of uncertainty estimation.
		Each panel shows the difference (ours minus probing) when training on one task
		and evaluating on a \emph{different task}, across all ordered task pairs.
		Diagonal entries correspond to same-task evaluation, while off-diagonal entries
		reflect true cross-task transfer.
		Positive values indicate improved performance of structured layer--layer
		signatures over probing under task shift.
	}
	\label{fig:e2}
\end{figure*}

Table~\ref{tab:e2} (Across-dataset) reports average performance across all task combinations, with detailed results for individual task pairs and models shown in Figure~\ref{fig:e2}. Diagonal entries correspond to same-task evaluation, while off-diagonal entries capture \emph{true cross-task generalization}, where both the data distribution and task semantics shift (e.g., sentiment classification $\rightarrow$ coreference resolution or question answering).


Both in aggregate and across models, we observe that our structured layer--layer \emph{signature maps} yield predominantly positive off-diagonal differences, indicating more robust transfer than probes trained on raw hidden states.
This suggests that signatures capture task-agnostic properties of cross-layer agreement that are predictive of correctness beyond the supervision signal of any single task, whereas probing features can overfit to task-specific cues.

\begin{table}[t]
	\caption{
		Comparison of \textbf{Signatures (ours)} versus \textbf{probing} in
		\textbf{within-dataset} and \textbf{across-dataset}
		settings.
		All values are mean \emph{absolute differences} (ours minus probing),
		reported in \emph{percentage points (pp)}.
		Higher is better; positive values are shown in bold.
	}
	\label{tab:e2}
	\centering
	\small
	\resizebox{\columnwidth}{!}{%
		\begin{tabular}{llcc}
			\toprule
			\textbf{Model} & \textbf{Metric} &
			\makecell{\textbf{Within-dataset}\\\textbf{(Diagonal, pp)}} &
			\makecell{\textbf{Across-dataset}\\\textbf{(Off-diagonal, pp)}} \\
			\midrule
			        
			\multirow{2}{*}{Llama-3.1-8B}
			  & AUPRC & $-1.80$         & $\mathbf{2.86}$  \\
			  & Brier & $\mathbf{4.88}$ & $\mathbf{21.02}$ \\
			\midrule
			        
			\multirow{2}{*}{Qwen3-14B}
			  & AUPRC & $-1.37$         & $\mathbf{0.95}$  \\
			  & Brier & $\mathbf{1.52}$ & $\mathbf{4.35}$  \\
			\midrule
			        
			\multirow{2}{*}{Mistral-7B-v0.3}
			  & AUPRC & $-1.72$         & $\mathbf{1.35}$  \\
			  & Brier & $\mathbf{3.38}$ & $\mathbf{4.28}$  \\
			\bottomrule
		\end{tabular}
	}
\end{table}

\subsection{E3: Quantization robustness (4-bit)}\label{sec:exp-e3}
Finally, we assess robustness to deployment-friendly 4-bit weight-only quantization. We train the estimator on full-precision activations and evaluate on activations from the same model loaded with 4-bit quantization. Table~\ref{tab:E3_combined} compares our method to probing on \textbf{Qwen3-14B-Instruct}. We observe that our method maintains strong \textbf{AUPRC} under quantization shift and yields consistently higher \textbf{Brier scores}, indicating resilience of the structured signals encoded by layer--layer divergences.

\begin{table*}[t]
	\centering
	\caption{Quantization shift (trained on full-precision, tested on 4-bit): Signature vs.\ Probing on \textbf{Qwen3-14B} (Instruct). \emph{Brier score is reported as $1-\mathrm{Brier\ loss}$, hence higher is better.}}
	\begin{tabular}{llcccccc}
		\toprule
		Metric & Method    & Math            & MMLU            & Movies          & WinoBias        & WinoGrande      & Mean            \\
		\midrule
		\multirow{2}{*}{AUPRC ($\uparrow$)}
		       & Signature & \textbf{0.9385} & \textbf{0.7303} & 0.9599          & 0.9151          & \textbf{0.5552} & \textbf{0.8198} \\
		       & Probing   & 0.9033          & 0.6594          & \textbf{0.9733} & \textbf{0.9171} & 0.5486          & 0.8004          \\
		\addlinespace
		\multirow{2}{*}{Brier score ($\uparrow$)} 
		       & Signature & \textbf{0.2445} & \textbf{0.4061} & 0.1837          & \textbf{0.3755} & \textbf{0.6406} & \textbf{0.3701} \\
		       & Probing   & 0.0438          & 0.1743          & \textbf{0.4542} & 0.3088          & 0.6029          & 0.3168          \\
		\bottomrule
	\end{tabular}
	\label{tab:E3_combined}
\end{table*}

\subsection{Performance and Representation Complexity Tradeoff}
Beyond raw predictive performance, it is also informative to consider the complexity of a UE method, as reflected by the size of the representation it operates on. Table~\ref{tab:dim-compare} provides an initial illustration of how quickly representation dimensionality can grow. We compare our approach to linear probing, ACT-ViT \citep{bar-shalom2025beyond}, and LOS-NET \citep{bar2025learning}. For ACT-ViT, we define the input dimensionality as $n_\text{layers}\times d_\text{hidden} \times T$, where $T$ denotes the number of tokens in the sequence. The input to LOS-NET has dimensionality $K \times T$ where $K$ is a method-specific constant introduced in \citet{bar2025learning}.

\begin{figure*}[t]
	\vspace{1em}
	\centering
	\begin{minipage}{0.49\textwidth}
		\centering
		\includegraphics[width=\linewidth]{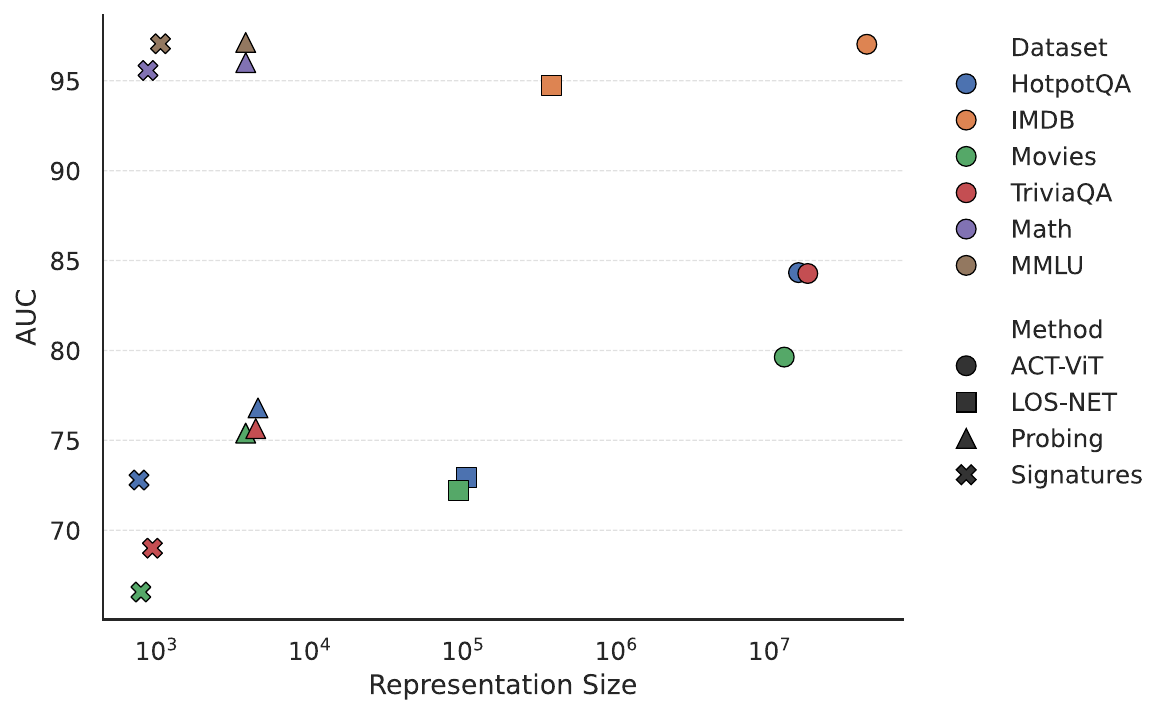}
		\vspace{-0.3em}
		\caption*{(a) Mistral-7B-v0.3-Instruct}
	\end{minipage}
	\hfill
	\begin{minipage}{0.49\textwidth}
		\centering
		\includegraphics[width=\linewidth]{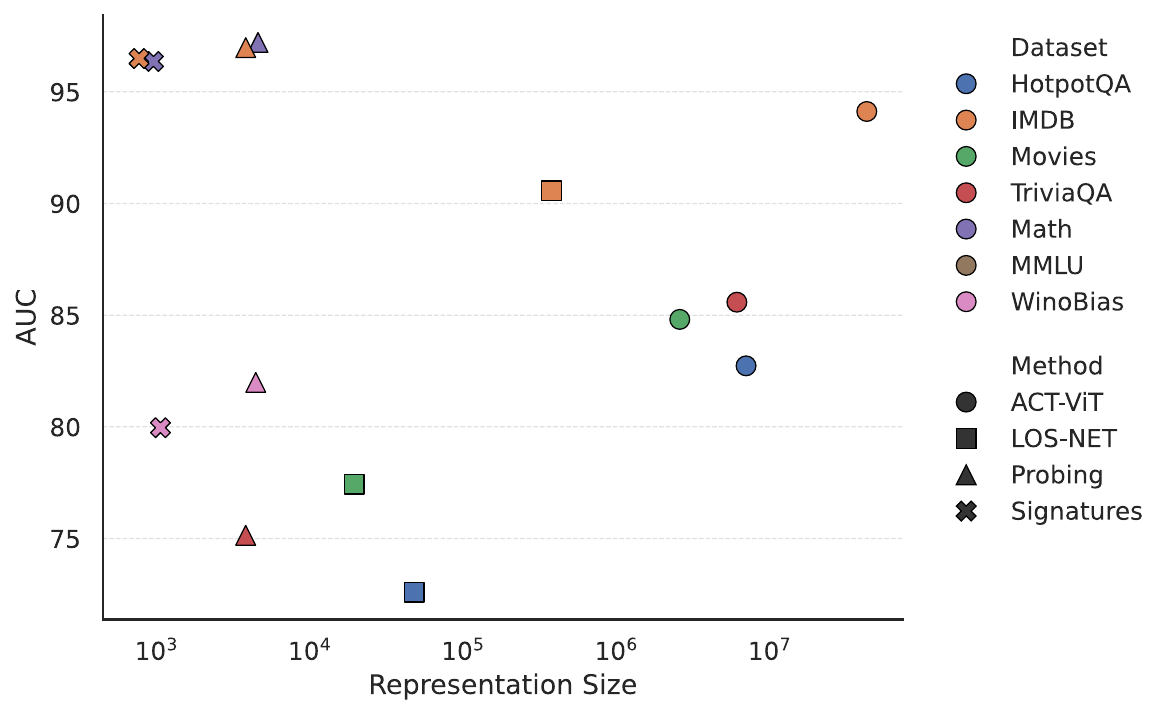}
		\vspace{-0.3em}
		\caption*{(b) Llama-3.1-8B}
	\end{minipage}
	
	\vspace{-0.5em}
	\caption{%
		Performance (AUC) as a function of UE method complexity, measured by the number of model parameters accessible to the method.
		Across both LLM backbones, our method achieves comparable performance to prior approaches while using several orders of magnitude fewer parameters.
	}
	\label{fig:auc-vs-complexity}
	\vspace{-1em}
\end{figure*}

Figure~\ref{fig:auc-vs-complexity} illustrates the trade-off between representation size and performance, measured using AUC. While AUC is not our primary evaluation metric---since it is less suitable than PRAUC for imbalanced settings---we report AUC here to ensure comparability with prior work. Importantly, the AUC results for LOS-NET and ACT-ViT were not recomputed in our experiments, but are taken directly from \citet{bar-shalom2025beyond} and \citet{bar2025learning}, respectively.

\begin{table}[H]
	\centering
	\caption{
		Representation dimension that uncertainty estimation methods can use.
		The representation used by our method (Signatures) is consistently the most compact.
		We assume $K=1{,}000$ and $T=100$.
		\newline
		$\dagger$~\citep{shalumov2023hero};
		$\ddagger$~\citep{openai2025};
		$\star$~\citep{Llama3Card2024};
		$\diamond$~\citep{microsoft2025phi4minitechnicalreportcompact};
		$\triangle$~\citep{Mistral7B2023};
		$\times$~\citep{qwen3technicalreport}.
	}
	\small
	\resizebox{\columnwidth}{!}{%
		\begin{tabular}{lrrrr}
			\toprule
			\textbf{Model} 
			& \makecell{$n_{\text{layers}}^2$ \\ {\small Ours}} 
			& \makecell{$d_{\text{hidden}}$ \\ {\small Probing}} 
			& \makecell{$n_{\text{layers}} \times d_{\text{hidden}} \times T$ \\ {\small ACT-ViT}} 
			& \makecell{$K \times T$ \\ {\small LOS-NET}} \\
			\midrule
			HeRo BERT\textsuperscript{$\dagger$}
			  & $\mathbf{144}$     & $3{,}072$ & $3{,}686{,}400$  & $100{,}000$ \\
			GPT-OSS 20B\textsuperscript{$\ddagger$}
			  & $\mathbf{576}$     & $2{,}880$ & $6{,}912{,}000$  & $100{,}000$ \\
			Llama 3.1 8B\textsuperscript{$\star$}
			  & $\mathbf{1{,}024}$ & $4{,}096$ & $13{,}107{,}200$ & $100{,}000$ \\
			Phi-4 Mini\textsuperscript{$\diamond$}
			  & $\mathbf{1{,}024}$ & $3{,}072$ & $9{,}830{,}400$  & $100{,}000$ \\
			Mistral 7B\textsuperscript{$\triangle$}
			  & $\mathbf{1{,}024}$ & $4{,}096$ & $13{,}107{,}200$ & $100{,}000$ \\
			Qwen3 14B \textsuperscript{$\times$}
			  & $\mathbf{1{,}600}$ & $5{,}120$ & $20{,}480{,}000$ & $100{,}000$ \\
			\bottomrule
		\end{tabular}
	}
	\label{tab:dim-compare}
\end{table}

\vspace{-1em}

\section{Conclusion and Future Work}\label{sec:discussion}
We proposed an inference-time uncertainty estimation method that learns global correctness predictors from local, instance-level evidence in the form of structured cross-layer divergence signatures. By transforming post-MLP activations into compact $L \times L$ directed KL signature maps and aggregating them with a lightweight classifier, the approach requires only a single forward pass, introduces no architectural modifications, and remains model-agnostic. This design explains its strong generalization: global decision rules are learned from local signals that reflect how information evolves across depth within individual sequences.

Conceptually, the method occupies a middle ground between Information Bottleneck (IB) analyses and probing. Unlike IB-style approaches, it does not rely on population-level mutual information estimates that are unavailable at inference time. Unlike probing, it avoids dependence on high-dimensional, opaque hidden states. The resulting signatures are information-theoretic, per-instance, and interpretable, while remaining practical and efficient. Empirically, this balance is reflected in performance: the method matches probing in-distribution despite using far fewer features, and consistently outperforms probing under distribution shift and weight-only quantization, achieving better AUPRC and Brier scores. These results indicate that structured cross-layer divergence patterns preserve transferable dynamics that are more robust than raw activations.

Beyond uncertainty estimation, the structured nature of the signatures enables new forms of interpretability. Feature attribution over divergence maps can identify which layer interactions most strongly influence correctness predictions, offering insight into how representations align or diverge across depth. Aggregating such maps across examples may reveal canonical ``agreement topologies,'' providing an inference-time analogue to IB-style narratives of fitting and compression. In this sense, the method serves both as a practical tool for calibration, abstention, and triage, and as a lens for studying internal dynamics of large language models.

\paragraph{Limitations and future work.}
The approach involves clear trade-offs. In strictly in-distribution settings, probes over full hidden states can sometimes match or slightly exceed performance, suggesting that the compressed signatures might discard fine-grained cues exploitable when train and test distributions coincide. The estimator is supervised and currently trained per task, making it sensitive to label quality and limiting direct cross-task or cross-domain transfer. Performance also depends on accurate selection of task-relevant tokens, and the compression inherent in the signatures precludes disentangling epistemic and aleatoric uncertainty. Future work should address these limitations through multitask and cross-domain training, semi-supervised or unsupervised variants, and richer token-level strategies that retain more signal while preserving structure. A deeper theoretical treatment connecting signature maps to broader information-theoretic quantities, as well as extensions to other architectures and generation settings, would further test the generality and scientific value of the framework.

\section{Reproducibility statement}
We follow the code infrastructure of \cite{orgad2025llms}, which allows for reproducible generation of the results in this paper. We refer to Appendix A in \cite{orgad2025llms} and follow the same reproducibility practices as there. All seeds, hyperparameters, data, and code are included as part of the supplementary material. 

\section{Impact Statement}
This paper aims to advance the field of machine learning by improving uncertainty estimation in large language models. More reliable uncertainty signals may support safer deployment practices such as abstention, calibration, and triage in downstream applications. The method operates post-hoc at inference time and does not introduce new model capabilities or access to sensitive data. We do not anticipate new societal risks arising from this work beyond those commonly associated with the deployment of large language models.
\bibliography{references}
\bibliographystyle{icml2026}
\vspace{-0.3em}
\appendix

\section{Datasets and Models}\label{app:data-models}

\subsection{Datasets}\label{app:datasets}
\textbf{TriviaQA} \citep{TriviaQA2017}. Open-domain QA with distant supervision from the web and Wikipedia. Questions are answered without supplying passages in our setup (parametric recall). Gold answers include alias sets; we treat a prediction as correct if, after standard normalization (lowercasing, punctuation/whitespace stripping), it matches any alias.

\textbf{HotpotQA} \citep{HotpotQA2018}. Multi-hop Wikipedia QA. We evaluate two regimes: \emph{no-context} (question only; stresses parametric multi-step reasoning) and \emph{with-context} (gold paragraphs provided; stresses grounded reasoning). Correctness uses exact-match against the gold short answer with standard normalization.

\textbf{Movies} (as constructed in \citealp{orgad2025llms}). Curated movie-knowledge QA with relatively narrow topical spread and short answers. We adopt the construction and splits of \citet{orgad2025llms}; correctness is string-match after normalization against their gold answers.

\textbf{WinoGrande} \citep{WinoGrande2020}. Adversarial Winograd-style pronoun resolution (cloze with two candidates). We format each item as a generation task and map the model output to one of the two candidates via normalized string matching; accuracy is the primary label.

\textbf{WinoBias} \citep{WinoBias2018}. Coreference resolution pairs designed to expose gender-bias confounds. We cast items as selecting the correct antecedent; predictions are mapped to the gold antecedent via normalized string matching; accuracy is reported. 

\textbf{IMDB} \citep{IMDB2011}. Binary sentiment classification over long movie reviews. We prompt for a discrete label (\texttt{positive}/\texttt{negative}) and score accuracy after normalization to canonical label tokens.

\textbf{Math (answerable subset)} (following \textsc{GSM8K} protocol; \citealp{GSM8K2021}). Grade-school math word problems. We restrict to answerable items per the protocol. Predictions are parsed for a final numeric/string answer and compared via normalized exact match.

\textbf{MMLU} \citep{MMLU2021}. 57 multiple-choice subjects spanning STEM, humanities, and social sciences. Each question has four options; we format it as question+options and score accuracy by matching the predicted option to the given ground-truth.

\section{Signature maps as a gateway to mechanistic analysis}
\label{sec:mechanistic-gateway}
Beyond predictive performance, a key motivation of our approach is that \emph{layer--layer signature maps admit structured interpretability}. Unlike probes over raw hidden states---whose features correspond to opaque coordinates in $\mathbb{R}^{d_{\text{hidden}}}$---each feature in our representation has a clear semantic meaning: a directed divergence between a specific pair of layers at a task-relevant token. This enables attribution not only to \emph{where} evidence arises in the network, but also to \emph{how information flows across depth}.

We analyze feature importance using the built-in explainability tools of gradient-boosted decision trees (LightGBM; \citealp{Ke2017LightGBM}), computing attributions based on split gains and second-order statistics (gradients and Hessians accumulated during training). These quantities measure the sensitivity of the loss to perturbations in each feature and provide a principled, model-internal ranking of predictive layer--layer interactions, though they are not causal.

Figures~\ref{fig:xai_qwen} and \ref{fig:xai_mistral} visualize these attributions for Qwen3-14B-Instruct and Mistral-7B-Instruct-v0.3. The models exhibit qualitatively different attribution profiles: for \textbf{Mistral}, importance decays monotonically with inter-layer distance, whereas \textbf{Qwen} shows a sparser pattern with influential interactions persisting across longer separations. We hypothesize that these differences reflect architectural or training-induced inductive biases, and their consistency across datasets suggests stable model-specific information flow rather than task-specific artifacts.

Although preliminary, this analysis suggests that layer--layer signature maps provide a practical bridge between black-box uncertainty estimation and mechanistic interpretability. By compressing high-dimensional activations into a structured interaction space, they enable attribution analyses expressed naturally in terms of layer dynamics. Aggregating attributions across inputs may further reveal canonical ``information flow signals'' associated with correct or incorrect behavior.

We do not claim a complete mechanistic account of LLM behavior. Rather, we view structured layer--layer signatures as a \emph{gateway} representation---expressive enough for accurate uncertainty estimation, yet constrained enough to support meaningful, architecture-level interpretation. Future work may combine this framework with causal interventions (e.g., layer ablations or representation patching) to move beyond correlational attribution toward mechanistic understanding.

\section{Comparing Different Information Scores}\label{sec:jsd-vs-kl}
As a preliminary ablation study, we compare Kullback-Leibler divergence vs Jensen-Shannon divergence. Results can be found in Table~\ref{tab:kld_vs_jsd_prauc}. The metrics are comparable, but $D_{\text{KL}}$ appears favorable; thus, we use it in our experiments.

\setcounter{table}{0}
\renewcommand{\thetable}{B\arabic{table}}

\begin{table}[H]
	\centering
	\caption{
		Comparison of PRAUC (test) between asymmetric KLD and symmetric JSD distance formulations across models and datasets.
		Best performance per model--dataset pair is bolded; 
	}
	\begin{tabular}{llcc}
		\toprule
		\textbf{Model} & \textbf{Dataset} & \textbf{$D_{\text{KL}}$} & \textbf{$D_{\text{JS}}$} \\
		\midrule
		\multirow{2}{*}{Llama-3.1-8B}
		               & IMDB             & \textbf{0.8752}          & 0.8719                   \\
		               & TriviaQA         & 0.4912                   & \textbf{0.4958}          \\
		\midrule
		\multirow{2}{*}{Mistral-7B-Instruct-v0.3}
		               & HotpotQA         & \textbf{0.7471}          & 0.7461                   \\
		               & TriviaQA         & \textbf{0.4604}          & 0.4578                   \\
		\midrule
		\multirow{3}{*}{Qwen3-14B}
		               & HotpotQA         & \textbf{0.8135}          & 0.8131                   \\
		               & IMDB             & 0.4614                   & \textbf{0.4906}          \\
		               & TriviaQA         & \textbf{0.5937}          & 0.5934                   \\
		\bottomrule
	\end{tabular}
	\label{tab:kld_vs_jsd_prauc}
\end{table}

\section{Comparison to LOS-NET and ACT-ViT}
A quantitative comparison between our method and LOS-NET \citet{bar2025learning} and ACT-ViT \citet{bar-shalom2025beyond} can be found in Tables~\ref{tab:signatures_vs_losnet} and~\ref{tab:signatures_vs_ACT_ViT}.
Across both comparisons, Signatures demonstrates consistently strong and often substantial gains, particularly in the most challenging reasoning-heavy settings. Notably, when trained and evaluated on HotpotQA, Signatures outperforms both LOS-NET (+13.83 pp) and ACT-ViT (+18.45 pp), indicating that it captures task-relevant structure more effectively under aligned supervision. It also achieves the best overall performance in 4/6 configurations against LOS-NET and remains competitive across heterogeneous transfer scenarios. These results suggest that Signatures provides a robust and effective modeling approach, delivering clear improvements in high-complexity regimes while maintaining stable cross-domain behavior.

\setcounter{table}{0}
\renewcommand{\thetable}{C\arabic{table}}
\begin{table}[t!]
	\centering
	\caption{Comparison of AUPRC (pp) between our method and LOS-NET on overlapping train--test pairs using Mistral-7B-Instruct-v0.3. Best result per pair is bolded.}
	\label{tab:signatures_vs_losnet}
	\begin{tabular}{llcc}
		\toprule
		Train & Method     & HotpotQA       & Movies         \\
		\midrule
		\multirow{2}{*}{HotpotQA}
		      & Signatures & \textbf{72.79} & \textbf{61.38} \\
		      & LOS-NET    & 58.96          & 58.98          \\
		\midrule
		\multirow{2}{*}{Movies}
		      & Signatures & 53.76          & \textbf{66.56} \\
		      & LOS-NET    & \textbf{58.96} & 58.98          \\
		\midrule
		\multirow{2}{*}{IMDB}
		      & Signatures & \textbf{59.21} & 51.72          \\
		      & LOS-NET    & 58.96          & \textbf{58.98} \\
		\bottomrule
	\end{tabular}
\end{table}

\begin{table}[h!]
	\centering
	\caption{Comparison of AUPRC (pp) between our method and ACT-ViT on overlapping train--test pairs using Mistral-7B-Instruct-v0.3. Best result per pair is bolded.}
	\label{tab:signatures_vs_ACT_ViT}
	\begin{tabular}{llcc}
		\toprule
		Train & Method     & HotpotQA       & TriviaQA       \\
		\midrule
		\multirow{2}{*}{HotpotQA}
		      & Signatures & \textbf{72.79} & 58.42          \\
		      & ACT-ViT    & 54.34          & \textbf{87.91} \\
		\midrule
		\multirow{2}{*}{TriviaQA}
		      & Signatures & \textbf{63.83} & 55.01          \\
		      & ACT-ViT    & 67.57          & \textbf{82.66} \\
		\midrule
		\multirow{2}{*}{IMDB}
		      & Signatures & 59.21          & \textbf{71.19} \\
		      & ACT-ViT    & \textbf{67.16} & 62.01          \\
		\bottomrule
	\end{tabular}
\end{table}

\newpage
\section{{Layer--Layer Interaction Patterns in Uncertainty Estimation}}
\setcounter{figure}{0}
\renewcommand{\thefigure}{E\arabic{figure}}
\begin{figure*}[t]
	\centering
		
	\begin{subfigure}[t]{0.26\linewidth}
		\centering
		\includegraphics[width=\linewidth]{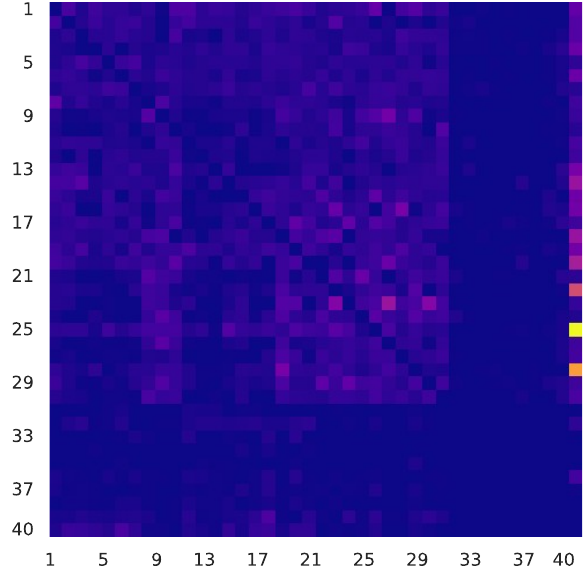}
		\caption{\textsc{WinoGrande}}
	\end{subfigure}\hspace{0.66cm}
	\begin{subfigure}[t]{0.26\linewidth}
		\centering
		\includegraphics[width=\linewidth]{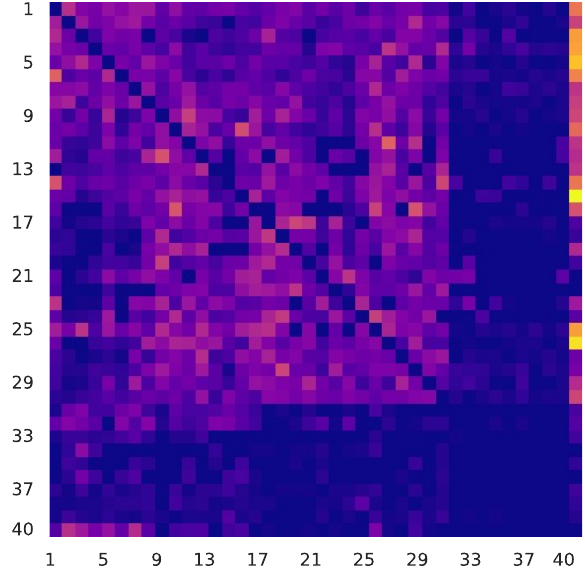}
		\caption{\textsc{TriviaQA}}
	\end{subfigure}\hspace{0.66cm}
	\begin{subfigure}[t]{0.26\linewidth}
		\centering
		\includegraphics[width=\linewidth]{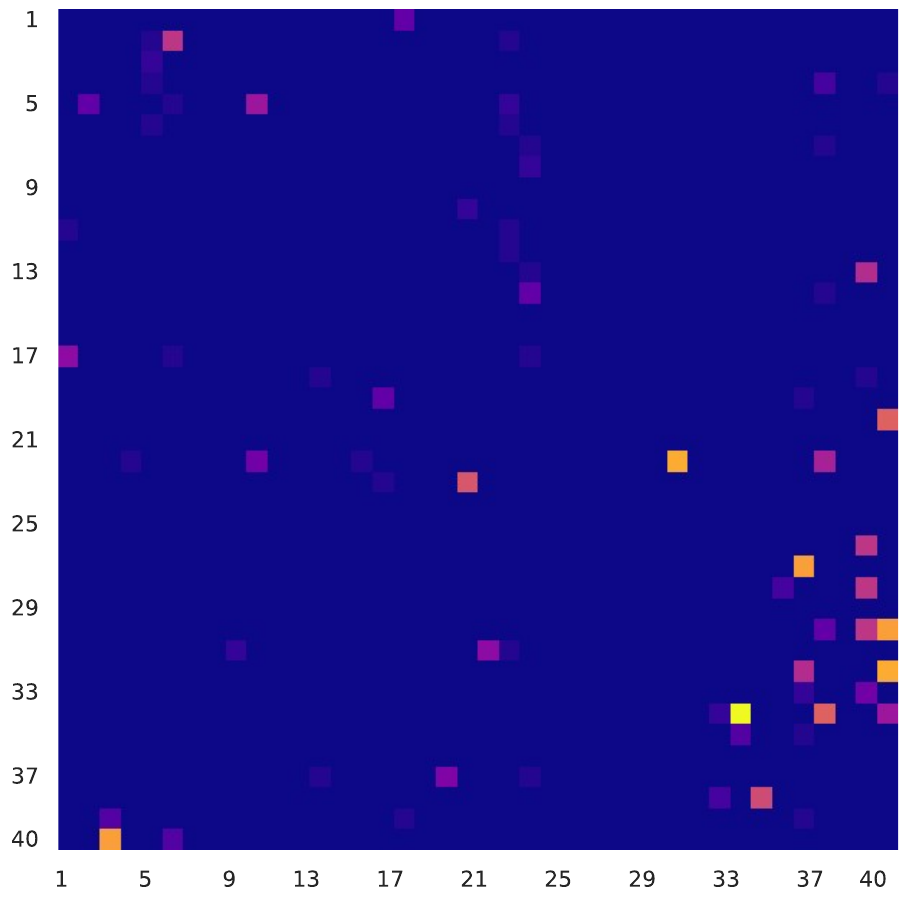}
		\caption{\textsc{MMLU}}
	\end{subfigure}
		
	\vspace{0.6em}
		
	\begin{subfigure}[t]{0.8\linewidth}
		\centering
		\includegraphics[width=\linewidth]{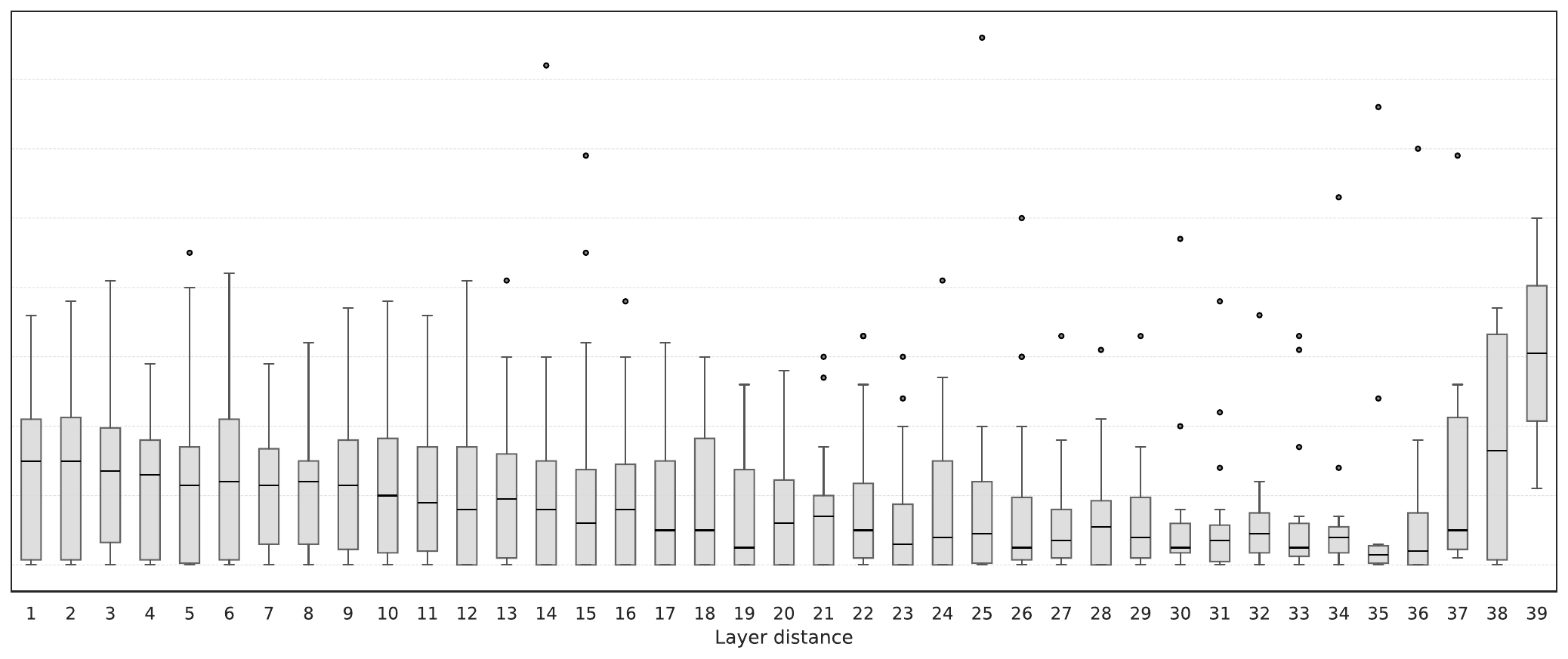}
		\caption{Aggregated importance vs.\ inter-layer distance}
		\label{fig:xai_qwen_distance}
	\end{subfigure}
	\vspace{-0.3em}
	\captionsetup{font=small}
	\caption{
		\textbf{Interpreting layer--layer interactions in Qwen3-14B-Instruct.}
		\textbf{Top:} Feature importance maps over layer--layer KL signatures for three datasets. Compared to Mistral, importance is distributed across a broader set of layer pairs.
		\textbf{Bottom:} Aggregation by inter-layer distance reveals a flatter profile, with influential interactions persisting across longer distances. This suggests that correctness-relevant information in Qwen is integrated over wider depth spans rather than being dominated by local layer interactions.
	}
	\label{fig:xai_qwen}
\end{figure*}

\begin{figure*}[t]
	\centering
		
	\begin{subfigure}[t]{0.26\linewidth}
		\centering
		\includegraphics[width=\linewidth]{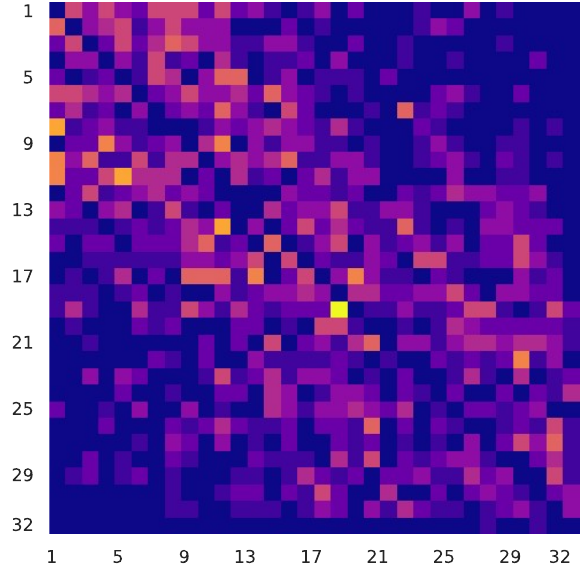}
		\caption{\textsc{WinoGrande}}
	\end{subfigure}\hspace{0.66cm}
	\begin{subfigure}[t]{0.26\linewidth}
		\centering
		\includegraphics[width=\linewidth]{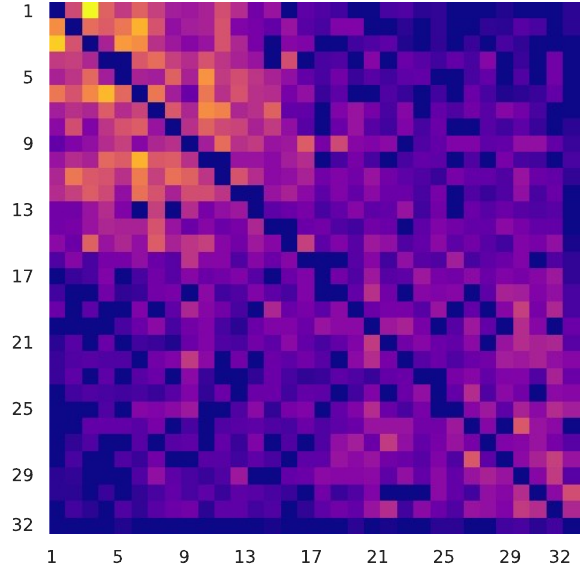}
		\caption{\textsc{Movies}}
	\end{subfigure}\hspace{0.66cm}
	\begin{subfigure}[t]{0.26\linewidth}
		\centering
		\includegraphics[width=\linewidth]{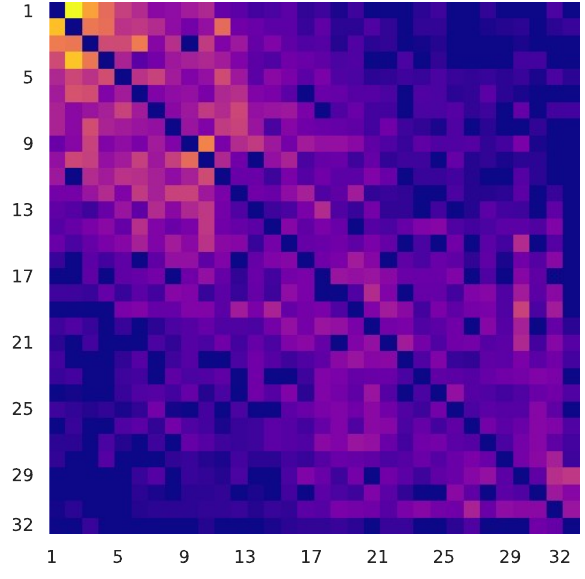}
		\caption{\textsc{TriviaQA}}
	\end{subfigure}
		
	\vspace{0.6em}
		
	\begin{subfigure}[t]{0.8\linewidth}
		\centering
		\includegraphics[width=\linewidth]{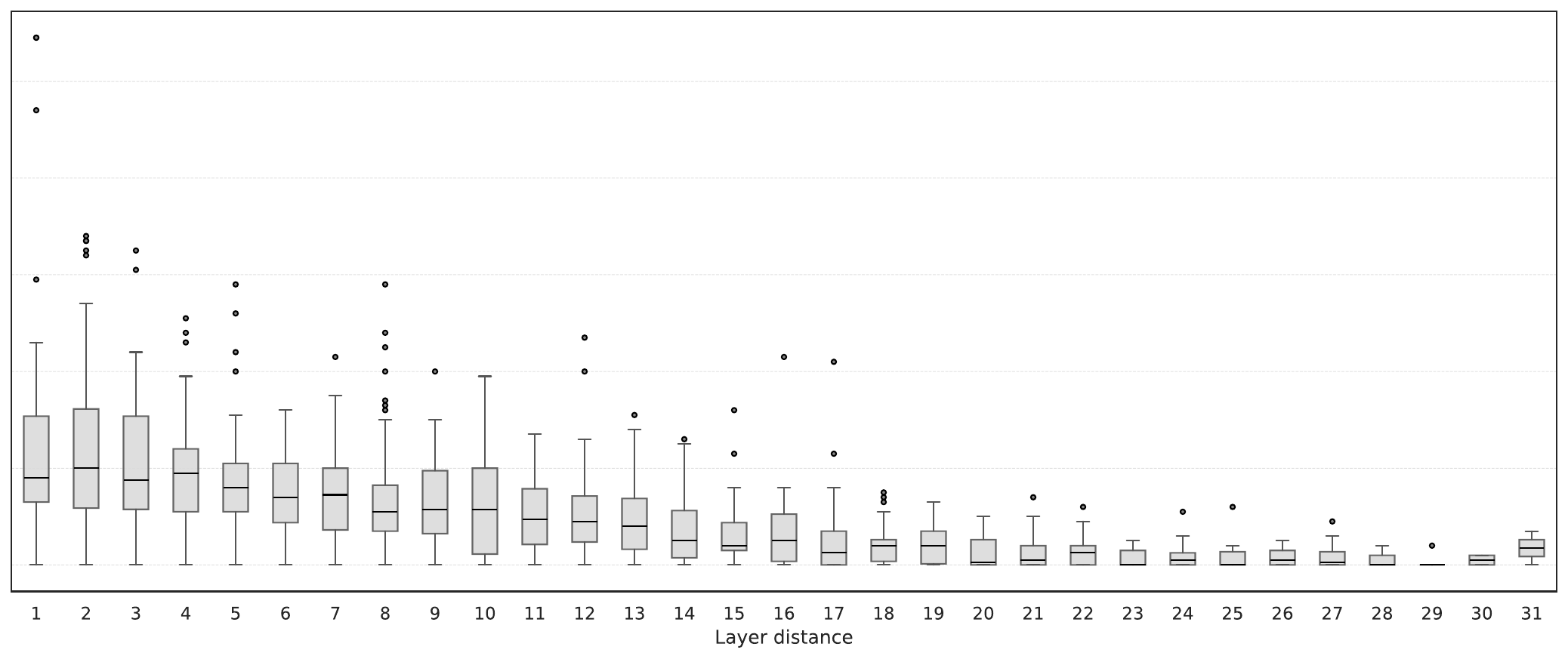}
		\caption{Aggregated importance vs.\ inter-layer distance}
		\label{fig:xai_mistral_distance}
	\end{subfigure}
	\captionsetup{font=small}
	\vspace{-0.3em}
	\caption{
		\textbf{Interpreting layer--layer interactions in Mistral-7B-Instruct-v0.3.}
		\textbf{Top:} LightGBM feature importance projected onto the $L{\times}L$ layer--layer signature maps for three datasets. Each cell corresponds to a directed KL divergence between a pair of layers; warmer colors indicate higher contribution to the correctness prediction.
		\textbf{Bottom:} Feature importance aggregated by inter-layer distance $|i-j|$, showing a clear decay as distance increases, consistent with predominantly local information refinement across depth.
	}
	\label{fig:xai_mistral}
\end{figure*}

\end{document}